\documentclass[
]{ceurart}

\usepackage{listings}
\usepackage{subcaption}
\begin{document}

\copyrightyear{2024}
\copyrightclause{Copyright for this paper by its authors.
  Use permitted under Creative Commons License Attribution 4.0
  International (CC BY 4.0).}

% \conference{Woodstock'22: Symposium on the irreproducible science,
%   June 07--11, 2022, Woodstock, NY}
\conference{The 2nd World Conference on eXplainable Artificial Intelligence,
17-19 July, 2024, Malta, Valletta}
  
\title{Interpretable Vital Sign Forecasting with Model Agnostic Attention Maps}

% \tnotetext[1]{You can use this document as the template for preparing your
%   publication. We recommend using the latest version of the ceurart style.}

\author[1]{Yuwei Liu}[%
email=yuwei.liu@spassmed.ca
]
% \cormark[1]
\address[1]{SpassMed Inc., Toronto, Ontario, Canada}
%108 College Street%

\author[1]{Chen Dan}[%
email=chen.dan@spassmed.ca
]

\author[1]{Anubhav Bhatti}[%
email=anubhav.bhatti@spassmed.ca
]

\author[1]{Bingjie Shen}[%
email=bingjie.shen@spassmed.ca
]

\author[1]{Divij Gupta}[%
email=divij.gupta@spassmed.ca
]

\author[1]{Suraj Parmar}[
email=suraj.parmar@spassmed.ca
]
\author[1]{San Lee}[%
email=sanlee@spassmed.ca
]

%Add email addresses of all and highlight corresponding author.

% \cortext[1]{Corresponding author.}
% \fntext[1]{These authors contributed equally.}

\begin{abstract}
Sepsis is a leading cause of mortality in intensive care units (ICUs), representing a substantial medical challenge. The complexity of analyzing diverse vital signs to predict sepsis further aggravates this issue. While deep learning techniques have been advanced for early sepsis prediction, their \textit{'black-box'} nature obscures the internal logic, impairing interpretability in critical settings like ICUs. This paper introduces a framework that combines a deep learning model with an attention mechanism that highlights the critical time steps in the forecasting process, thus improving model interpretability and supporting clinical decision-making. We show that the attention mechanism could be adapted to various black box time series forecasting models such as N-HiTS and N-BEATS. Our method preserves the accuracy of conventional deep learning models while enhancing interpretability through attention-weight-generated heatmaps. We evaluated our model on the eICU-CRD dataset, focusing on forecasting vital signs for sepsis patients. We assessed its performance using mean squared error (MSE) and dynamic time warping (DTW) metrics. We explored the attention maps of N-HiTS and N-BEATS, examining the differences in their performance and identifying crucial factors influencing vital sign forecasting.
\end{abstract}

\begin{keywords}
  Time Series Forecasting \sep
  Deep Learning \sep
  Interpretable Machine Learning \sep
  Attention Map \sep
  Vital Signs \sep
  Sepsis \sep
  Explainable AI
\end{keywords}

\maketitle

\section{Introduction}
Sepsis is a life-threatening condition that occurs when the immune system of the body responds incorrectly to an infection and causes rapid organ dysfunction and failure \cite{gul2017changing}. A meta-analysis conducted on articles published in PubMed and the Cochrane Database revealed that the average 30-day mortality rate for sepsis was 24.4\%, and the average 90-day mortality rate was 32.2\% between 2009 and 2019 \cite{bauer2020mortality}. While sepsis has been acknowledged for a long time, its clinical definition did not emerge until the late 20$^{th}$ century \cite{gotts2016sepsis}. In 1991, a consensus conference posited that sepsis arises from the individual's inflammatory response to infection, marked by systemic inflammatory response syndrome (SIRS), emphasizing the human response to invading organisms. This syndrome is characterized by variations in temperature, heart rate (HR), respiratory rate (RR), blood pressure (BP), and white blood cell (WBC) count \cite{vincent2013sepsis}. In 2016, the definition of sepsis was revised to multiple organ dysfunction syndrome (MODS) \cite{cheng2020critical}. Systolic blood pressure (SBP) and RR abnormalities indicate organ dysfunction \cite{singer2016third}. Thus, creating precise models for forecasting vital signs becomes essential in predicting sepsis \cite{behinaein2021transformer}. Accurate vital sign predictions can promptly aid clinicians in identifying and intervening in sepsis cases, potentially saving lives and improving the intensive care unit (ICU) patient outcomes.

The growth in explainable artificial intelligence (XAI) research is mainly attributed to the rapid growth in the popularity of deep learning with widespread healthcare applications. However, most models developed using these technologies are considered \textit{'black-boxes'} by experts due to their intricate, non-linear structures that are challenging for non-experts to understand \cite{vilone2021notions}. The proposed research contributes to the following two aspects: \textbf{(1)} Adding an attention mechanism to show the relationship between input time steps and forecasted results; \textbf{(2)} Providing analysis and interpretation of the findings derived from the attention map.

\subsection{Literature Review}
In recent years, the significance of model explainability has been widely recognized, leading to the integration of an increasing number of explainable methods into data-driven models \cite{longo2020explainable}. Prior research has demonstrated the development of deep learning neural networks incorporating attention mechanisms, resulting in interpretable models with strong performance within the medical field. Kaji et al. demonstrated that integrating an attention mechanism into the LSTM network, trained with Electronic Health Record (EHR) data, not only improves the daily sepsis onset prediction's Area Under the Receiver Operating Characteristic Curve (AUROC) score to 0.876 but also highlights critical time points for prediction \cite{kaji2019attention}. An attention-based gated recurrent unit (GRU) was developed by Shickel et al. Self-attention was applied to focus on significant time steps when predicting in-hospital mortality \cite{shickel2019deepsofa}. Choi et al. proposed reverse time attention (RETAIN), processing EHR data in reverse order, achieving an Area Under the ROC Curve (AUC) of 0.87 in heart failure prediction. It adds interpretability using a two-level neural attention model \cite{choi2016retain}.

While previous XAI research integrating deep learning models with interpretable modules has excelled in time series classification, attention mechanisms in interpretable time series forecasting remain underexplored. Our approach aims to explore attention mechanism interpretability in time series forecasting.

\section{Method}

In this section, we begin by detailing the information of the eICU Collaborative Research Database (eICU-CRD) \cite{pollard2018eicu}, followed by an outline of the composition of our input data. Subsequently, we dive into the specifics of the attention mechanism and the frameworks of our forecasting models. %This is succeeded by a comprehensive explanation of the results obtained from our analysis.%

\subsection{Dataset Description and Data Preprocessing}
The eICU-CRD data is a publicly accessible repository containing data from over 200,000 ICU admissions across 208 hospitals in the United States between 2014 and 2015 \cite{pollard2018eicu}. This comprehensive dataset comprises diverse patient information, including demographics, diagnoses, medications, and laboratory results. Our research focuses on the 'diagnosis' and 'vitalAperiodic' tables, from which we extract dynamic physiological data such as temperature, HR, and BP at 5-minute intervals. The core of our study revolves around forecasting two crucial dynamic variables: HR and mean blood pressure (MBP), derived from SBP and diastolic blood pressure (DBP) measurements. %Maybe mention some rationale for doing so? maybe 'Following the works of \cite{Anubhav paper 1, Anubhav paper 2}, we create ..'
Following the works of \cite{bhatti2023interpreting, o2020characterizing}, we create one or more groups within a 9-hour time window for each patient to predict vital signs for the subsequent 3 hours based on the preceding 6 hours of data. Data preprocessing involves imputing missing values, filtering outliers, and scaling using domain-specific knowledge. Clinically reasonable boundaries for each critical vital sign were set using this specialized knowledge: HR ranged from 0 to 300 bpm, MBP from 0 to 190 mmHg, and RR from 0 to 100 bpm \cite{parmar2024extending}.

\subsection{Experiment Setup}
The dataset is divided into training, validation, and test sets in an 80:10:10 ratio. Within these intervals, the initial 6 hours consist of 72 time steps, while the subsequent 3 hours encompass 36 time points. The forecasting model integrates either HR alone or HR combined with RR as covariates to forecast MBP or conversely. Training of the model occurs over the first 72 time steps, followed by predictions for the remaining 36 time steps. Ultimately, model performance is assessed through Mean Squared Error (MSE) and Dynamic Time Warping (DTW) evaluations.

\subsection{Deep Learning Forecasting Model}

\begin{figure}[b]
     \centering
     \fcolorbox{black}{white}{\includegraphics[width=\textwidth]{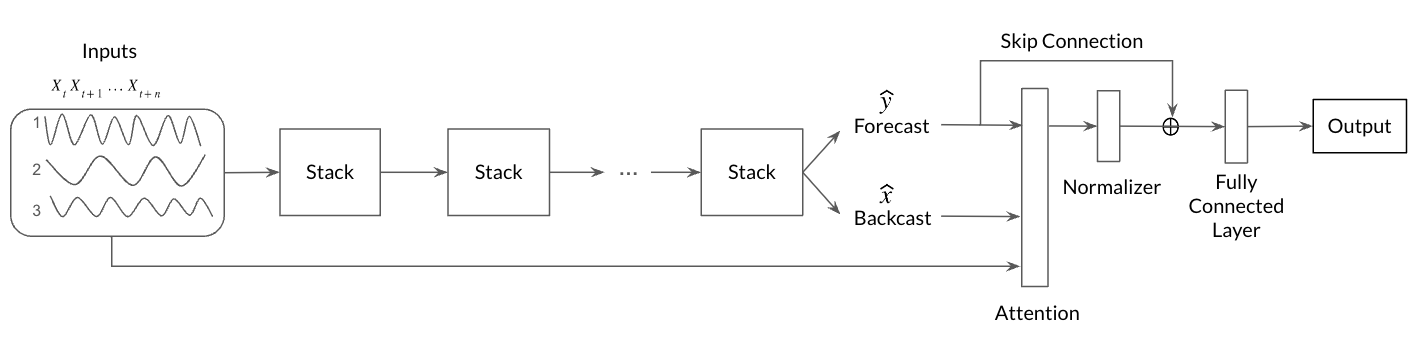}}
    \caption{Structure of our implementation. Adding the attention layer at the top of stacks, getting the results from the output.}
    \label{implementation}
\end{figure}

Based on the forecasting performance of the N-HiTS and N-BEATS model \cite{challu2023nhits,oreshkin2020nbeats,bhatti2023vital}, as well as the idea proposed by Pantiskas et al. \cite{pantiskas2020interpretable} we aim to address their inherent lack of interpretability and understand why the model has different performances. To achieve this, we implemented an attention mechanism that can be applied to the N-HiTS and N-BEATS architecture, which may also be applied to other black-box deep learning models. The N-HiTS and N-BEATS model consists of a series of stacks, each responsible for learning residual values from the preceding stack. 

Within each stack are blocks comprising several fully connected layers, which generate backward ($\theta^{b}_{l}$) and forward ($\theta^{f}_{l}$) expansion coefficients according to Equation \ref{nbeats:linear} %\cite{challu2023nhits}
, where $h_{l, 4}$ represents the output of the fourth fully connected layer in the basic block, and $Linear$ denotes a linear projection layer \cite{challu2023nhits}:
\begin{equation}
\label{nbeats:linear}
    \theta^{b}_{l} = \text{Linear}^{b}_{l}(h_{l, 4}), \thickspace \thickspace
    \theta^{f}_{l} = \text{Linear}^{f}_{l}(h_{l, 4}),
\end{equation} 

Additionally, each block includes backward ($g^{b}_l$) and forward ($g^{f}_{l}$) basis layers that produce backcast and forecast outputs as per Equation \ref{nbeats:basis}, where $\widehat{y}_{l}$ and $\widehat{x}_{l}$ denote forecast and backcast outputs, respectively:

\begin{equation}\label{nbeats:basis}
        \widehat{y}_{l} = \sum_{i=1}^{\text{dim}(\theta^{f}_{l})} \theta^{f}_{l,i}\text{v}^{f}_{l,i}, \thickspace \thickspace
        \widehat{x}_{l} = \sum_{i=1}^{\text{dim}(\theta^{b}_{l})} \theta^{b}_{l,i}\text{v}^{b}_{l,i}.
\end{equation}

Here, $v^{f}_{l,i}$ and $v^{b}_{l,i}$ represent forecast and backcast basis vectors. Notably, for N-HiTS, it has a max-pooling layer (Equation \ref{nhits:maxpool}) before passing the values to the fully connected layer, which is applied to enable multi-rate signal sampling for the $l^{th}$ basic block \cite{challu2023nhits}:

\begin{equation}\label{nhits:maxpool}
    y^{(p)}_{t - L:t, l} = \text{MaxPool}\left(y_{t - L:t, l}, k_{l}\right),
\end{equation}
where $k_{l}$ is the kernel size of the MaxPool layer.

Subsequently, inspired by Pantiskas et al. \cite{pantiskas2020interpretable} idea, we introduced an attention mechanism to explore the relationship between learned information and original inputs after obtaining the residuals from the final stack. The forecasted result is utilized to construct the Query (Q), while the original input forms the basis for the Value (V) and Key (K) \cite{pantiskas2020interpretable}. The resulting output is computed as follows:

\begin{equation}
    O^{N*H} = D \cdot V = softmax(\frac{QK^T}{\sqrt{L}})V 
\end{equation}
\begin{equation}
    K^{N*1*L} = I^{N*1*L} \cdot W^{N*L*L}_K + b^{N*1*L}_K 
\end{equation}

\begin{equation}
    V^{N*1*L} = I^{N*1*L} \cdot W^{N*L*L}_V
 \end{equation}

and $N$ is the number of input multi time seires, $H$ is the forecasting horizon length, and $L$ is the history input horizon. As shown in Figure \ref{implementation}, after the attention layer, a normalizer is applied, and skip connections are employed to mitigate the vanishing gradient issue. Finally, a fully connected layer is utilized to generate the forecasted results.

\subsection{Interpretable Attention Map}
To illustrate the attention map for a specific item, we computed \cite{pantiskas2020interpretable}:

\begin{equation}
A^{H*L*N} = D^{H*L} \cdot abs(W^{N*L*L}_v)^T 
\end{equation}

Here, $A^{H*L*N}$ denotes the attention map, where $A^{H*L}_i$ represents the $i^{th}$ series in the multivariate time series. Each row $j$ in $A^{H*L}_i$ signifies the relationship between the $j^{th}$ forecasted data point and the historical input of length $L$.

This computation enables the visualization of how the model attends to different historical inputs when forecasting specific data points across the multivariate time series.

\section{Results and Discussion}
%Some lines to start the section.
\subsection{Forecasting Benchmarks}

\begin{table*}
  \caption{Performance of forecasting models on forecasting MBP and HR. Here, covariates (W C) for MBP are HR \& RR, and covariates for HR are MBP \& RR. $^*$The MSE values are scaled by $1e^{-4}$ for better representation. $^\dagger$The DTW values are scaled by $1e^{-3}$ for better representation.
  }
  \label{tab:results}
  \begin{tabular}{ccccccl}
    \toprule
    {\textbf{Models}} & {\textbf{Cov.}} & {\textbf{MBP (MSE*)}} & {\textbf{MBP (DTW$^\dagger$)}} &
    {\textbf{HR (MSE*)}} & {\textbf{HR (DTW$^\dagger$)}}\\
    \midrule
    {Persistence \cite{bhatti2023vital}} & - &  24.55 &  34.50 &  7.35 &  17.52 \\
    {N-HiTS \cite{bhatti2023vital}} & W C &  18.46 &  18.70 &  7.37 &  13.12 \\
     {N-HiTS \cite{bhatti2023vital}} & W/o C &  \textbf{18.02} &  20.46 &  7.22 &  13.97  \\
    {N-BEATS \cite{bhatti2023vital}} & W C &  19.79 &  19.37 &  8.73 &  14.36 \\
    {N-BEATS \cite{bhatti2023vital}} & W/o C &  18.52 &  \textbf{17.63} &  7.48 &  \textbf{10.71} \\
    {TFT \cite{bhatti2023vital}} & W C &  18.89 &  25.93 &  7.71 &  16.12 \\
    {TFT \cite{bhatti2023vital}} & W/o C &   19.45 &  25.65 &  8.12 &  16.65 \\
    {N-BEATS with Attention} & W C & 21.86 & 21.07 & 8.04 & 14.32 \\ 
    {N-BEATS with Attention} & W/o C &  18.71 & 18.03  & 8.40  &  11.33 \\ 
    {N-HiTS with Attention} & W C &  18.78 &  20.44 &  7.24 &  13.32 \\
    {N-HiTS with Attention}  & W/o C &  19.73 &  20.42 &  \textbf{6.97} &   12.24 \\ 
     \bottomrule
    \end{tabular} 
\end{table*}

\begin{figure}
     \centering  
     \begin{subfigure}[b]{0.8\textwidth}
         \centering
         \includegraphics[width=\textwidth]{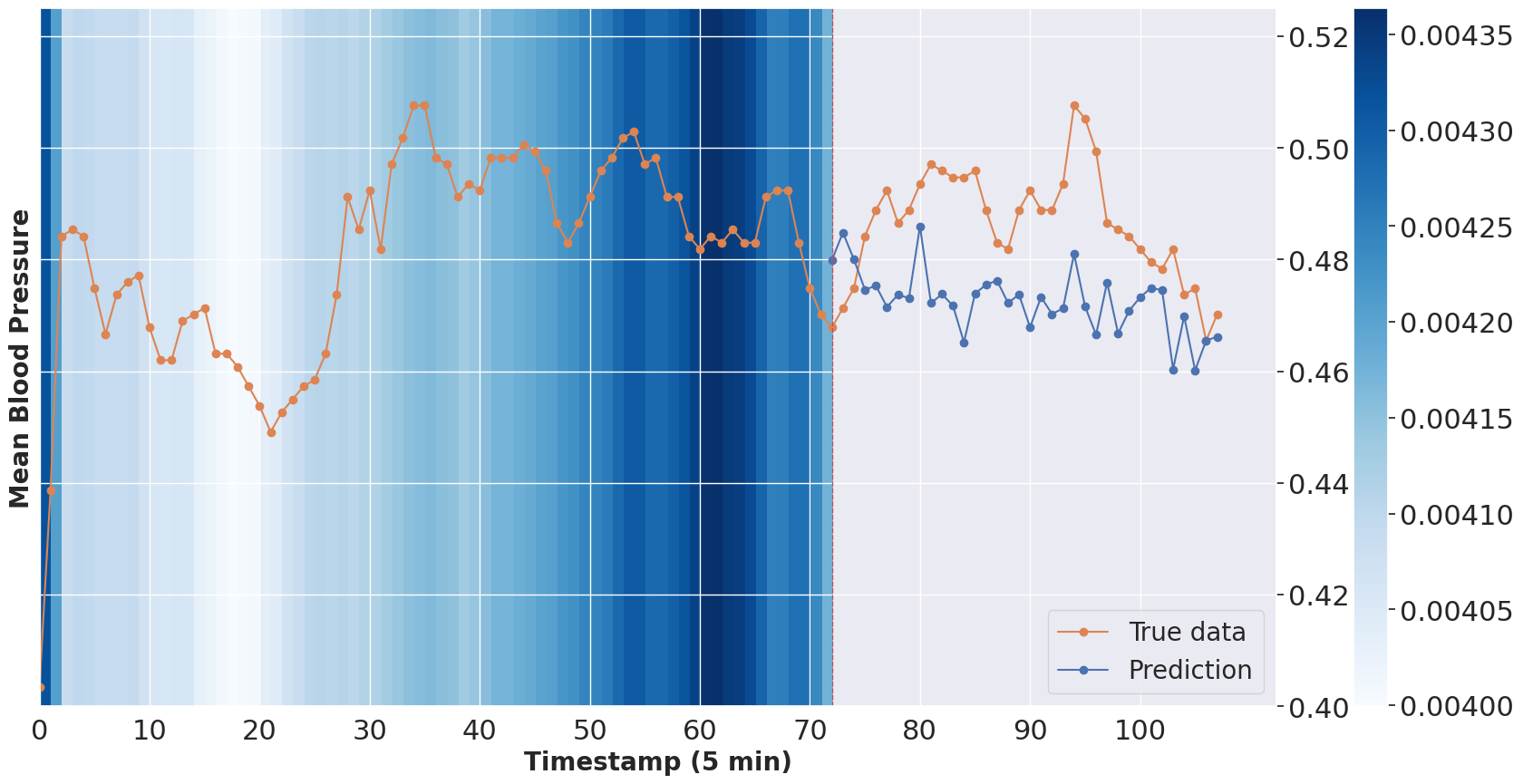}
         \caption{N-HiTS Attention distribution}
         \label{fig:attentions}
     \end{subfigure}
     \begin{subfigure}[b]{0.8\textwidth}
         \centering
         \includegraphics[width=\textwidth]{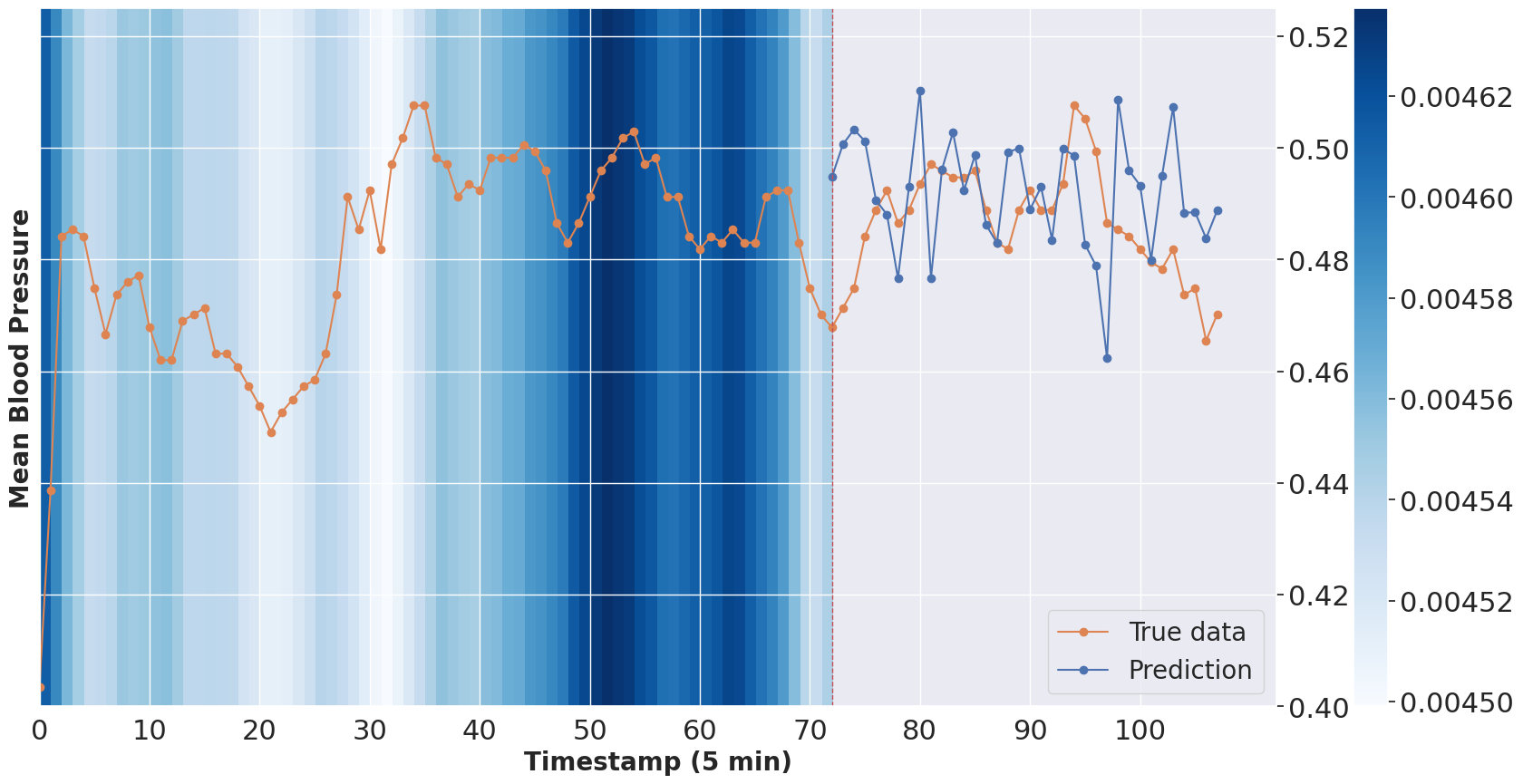}
         \caption{N-BEATS Attention distribution.}
          \label{fig:results_nbeats}
     \end{subfigure}
     % \hfill
       \caption{N-HiTS \& N-BEATS with attention using covariates to forecast MBP after minmax filter.}
        \label{fig:results}
     
\end{figure}

 \begin{figure}
     \centering
     \includegraphics[width=0.72\textwidth]{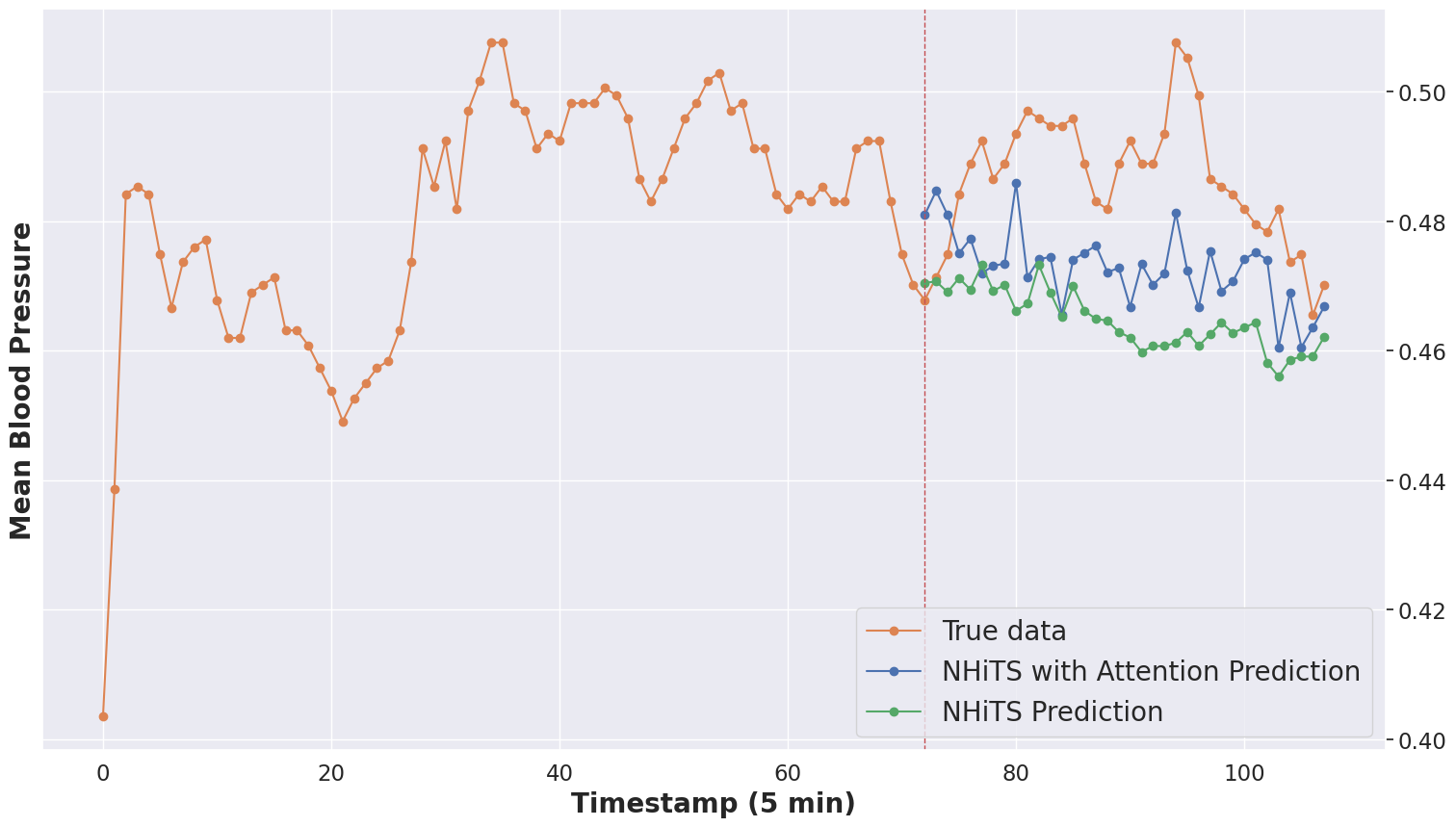}
     \caption{N-HiTS forecasting results with attention using covariates after minmax filter}
     \label{fig:forecasting}
 \end{figure}
      
Here, table \ref{tab:results} shows the results using different deep learning time series forecasting models. We compared N-HiTS \cite{challu2023nhits}, N-BEATS \cite{oreshkin2020nbeats}, Temporal Fusion Transformer (TFT) \cite{lim2021temporal}, which are computed by Bhatti et al. \cite{bhatti2023vital} using MSE and DTW as the evaluation metrics.

The results indicate that the N-HiTS model, both with and without an attention mechanism, consistently outperforms other models across MBP and HR predictions when considering MSE. Similarly, the N-BEATS model also performs well both with and without attention mechanisms.

Furthermore, the TFT model demonstrates competitive performance, especially when considering MSE. But in the previous paper by Bhatti et al. \cite{bhatti2023vital}, the forecasting result of TFT is relatively smooth and doesn't show fluctuations. 

In conclusion, the N-HiTS model, when augmented with an attention mechanism, emerges as a robust choice for forecasting MBP and HR, showcasing its efficacy in capturing complex temporal patterns. However, further exploration and experimentation are warranted to optimize model performance, particularly regarding temporal alignment and covariate incorporation.

\subsection{Interpretability Analysis}

In the heatmap provided (Fig \ref{fig:attentions}, Fig \ref{fig:results_nbeats}), darker colors indicate higher attention weights at specific time points, which correspondingly have a greater influence on prediction outcomes. Conversely, lighter colors suggest a lesser impact. The "N-HiTS + Attention"  in Fig \ref{fig:attentions} demonstrates that areas after the 20$^{th}$ time point exhibit darker shades compared to earlier sections. Notably, significant changes or peaks at certain points (like the 35$^{th}$, 54$^{th}$, and 63$^{rd}$ points) increasingly darken, highlighting their crucial role in shaping the prediction. This pattern suggests that N-HiTS places a stronger emphasis on data after the 20$^{th}$ points, effectively capturing both data fluctuations and overall trends. As a result, the predictions closely align with the actual data and accurately reflect downward trends.

On the other hand, the predictions from N-BEATS do not closely follow the downward trend of the actual data and display considerable fluctuation. This model’s attention map reveals that N-BEATS in Fig \ref{fig:results_nbeats} assigns larger weights to almost every rise and fall (such as at the 3$^{rd}$, 10$^{th}$, and 29$^{th}$ points), but without considering if it’s worth to focus on the trend, which contributes to less effective information capture. Moreover, it appears that N-BEATS prioritizes data from the initial 1-2 hours more than N-BEATS, contributing to less stable prediction outcomes.

Both models indicate that the initial 1-3 hours are crucial for prediction, suggesting that medical staff should focus on interventions during this period. Significant changes occurring up to three hours prior also substantially impact the predictions.

\section{Conclusion}
% In this paper, we presented an innovative interpretable time series forecasting algorithm that combines N-Hits with an attention mechanism. This approach allows us to observe how the deep learning algorithm assigns importance to inputs while generating each step of its output in a transparent manner. Upon applying this advanced architecture to the eICU dataset, our findings demonstrate that the attention mechanism can enhance interpretability in deep learning time series forecasting models with minimal reduction or even no change in accuracy. By visualizing the forecasted outcomes alongside the attention distributions, users gain a deeper understanding of the model's decision-making rationale and can better interpret the forecasted results in the context of the underlying data dynamics. This interpretability aspect is crucial, particularly in domains where transparency and explainability are paramount, such as reseaarching on sepsis and septic shock area.
%%%%%%
In this paper, we presented an interpretable time series forecasting algorithm that combines black-box deep learning models (N-HiTS \& NBEATS) with a general attention mechanism. This approach allows us to observe how the deep learning algorithm assigns importance to inputs while transparently generating each step of its output. Upon applying this advanced architecture to the eICU-CRD dataset, our findings demonstrate that the attention mechanism can enhance interpretability in deep learning time series forecasting models with minimal reduction or even no change in accuracy. By visualizing attention distributions, clinicians can identify which vital signs and historical data points are most influential in predicting sepsis. Furthermore, our model-agnostic attention mechanism is applicable to various deep learning forecasting models.

%% Define the bibliography file to be used
\bibliography{main}


% The sources for the ceur-art style are available via
% \begin{itemize}
% \item \href{https://github.com/yamadharma/ceurart}{GitHub},
% % \item \href{https://www.overleaf.com/project/5e76702c4acae70001d3bc87}{Overleaf},
% \item
%   \href{https://www.overleaf.com/latex/templates/template-for-submissions-to-ceur-workshop-proceedings-ceur-ws-dot-org/pkfscdkgkhcq}{Overleaf
%     template}.
% \end{itemize}

\end{document}